\documentclass{article} 
\usepackage[a4paper,left=1.2in,right=1.2in,top=1.2in,bottom=1.2in]{geometry}

\usepackage[utf8]{inputenc} 
\usepackage[T1]{fontenc}    
\usepackage{hyperref}       
\usepackage{url}            
\usepackage{booktabs}       
\usepackage{nicefrac}       
\usepackage{microtype}      
\usepackage{pifont}

\usepackage[sc]{mathpazo}
\usepackage{enumitem}

\usepackage{wrapfig}
\usepackage{changepage}
\usepackage{amsmath}
\usepackage{amsthm}
\usepackage{amssymb}
\usepackage{amsfonts,eucal,amsbsy}
\usepackage{mathtools}
\usepackage{multirow}
\usepackage[normalem]{ulem}
\usepackage{natbib}
\usepackage[usenames,dvipsnames]{xcolor}

\usepackage{subfigure}

\newcommand{\ignore}[1]{}
\newcommand{\vect}[1]{\mathbf{#1}}

\newcommand{\dycomment}[1]{\textcolor{blue}{\bf \small [#1 --DY]}}

\newcommand{\cjd}[1]{\textcolor{cyan}{\bf \small [#1 -cjd]}}

\usepackage{titling}
\title{Learning and Evaluating \\ General Linguistic Intelligence}
\author{\textbf{Dani Yogatama, Cyprien de Masson d'Autume, Jerome Connor,} \\
\textbf{Tomas Kocisky, Mike Chrzanowski, Lingpeng Kong,} \\
\textbf{Angeliki Lazaridou, Wang Ling, Lei Yu, Chris Dyer, Phil Blunsom}\\
DeepMind, London, United Kingdom\\
\texttt{\{dyogatama,cyprien,jeromeconnor,tkocisky,chrzanowskim\}@google.com}\\
\texttt{\{lingpenk,angeliki,lingwang,leiyu,cdyer,pblunsom\}@google.com}
}
\date{}

\begin{document}

\maketitle

\begin{abstract}
We define general linguistic intelligence as the ability
to reuse previously
acquired knowledge about a language's lexicon, syntax, semantics, and pragmatic conventions to adapt 
to new tasks quickly.
Using this definition, we analyze state-of-the-art natural 
language understanding models and conduct
an extensive empirical investigation to 
evaluate them against these criteria through a series of experiments that assess the task-independence of the knowledge being acquired by the learning process.
In addition to task performance, we propose a new
evaluation metric based on an online encoding
of the test data that quantifies how quickly an existing 
agent (model) learns a new task.
Our results show that while the 
field has made impressive progress
in terms of model architectures
that generalize to many tasks,
these models still require 
a lot of in-domain training examples 
(e.g., for fine tuning, training task-specific modules),
and are prone to catastrophic forgetting.
Moreover, we find that far from solving 
general tasks (e.g., document question answering), 
our models are overfitting to the quirks 
of particular datasets (e.g., SQuAD).
We discuss missing components and conjecture on how to make
progress toward general linguistic intelligence.

\end{abstract}

\section{Introduction}
Advances in deep learning techniques (e.g., attention mechanisms, 
memory modules, and architecture search) 
have considerably improved
natural language processing (NLP) models on 
many important tasks.
For example, machine performance on both Chinese--English machine translation 
and document question answering on 
the Stanford question answering dataset 
(SQuAD; Rajpurkar et al., 2016) \nocite{squad}
has been claimed to have surpassed human levels \citep{chinesemt,bert}.
While the tasks that initiated learning-based 
NLP models were motivated by external demands
and are important applications in their own right 
(e.g., machine translation, question answering,
automatic speech recognition, text to speech, etc.),
there is a marked and troubling tendency for 
recent datasets to be set up to be easy to 
solve with little in the way of 
\emph{generalization} or \emph{abstraction}; for instance, 
ever larger datasets created by crowd-sourcing processes that may not well approximate the natural distributions they are intended to span, although there are some notable counter-examples~\citep{naturalq}. 
When there exist multiple datasets 
that are representative of the exact same 
task from different domains (e.g., various question answering datasets), 
we rarely evaluate on all of them. 
This state of affairs
promotes development of models 
that only work well for a specific purpose,
overestimates our success at having solved the 
general task,
fails to reward sample efficient generalization
that requires the ability to discover 
and make use of rich linguistic structures,
and ultimately limits progress.

Despite these methodological difficulties, recent breakthroughs demonstrate that models that are pretrained 
on a large unlabeled corpus perform well
on many language understanding tasks with minimal 
modifications \citep{openai, elmo, bert}.
These models are trained 
on a large corpus in an unsupervised fashion
on tasks such as language modeling 
(predicting the next word given a context), masked language modeling
(predicting a missing word in a sentence from the context), 
and next sentence predictions 
(predicting whether two sentences are consecutive sentences).
The models can then be used on various NLP tasks
by adding an extra task-specific final layer 
and fine tuning on supervised data for the task
of interest.
Variants of these models achieve impressive results on 
the GLUE benchmark (sentence and sentence pair classification tasks; 
Wang et al., 2018) \nocite{glue} 
and the SQuAD question answering dataset,
showing the clear benefit of, and significant progress on,
transfer learning in NLP.

Another promising avenue is to train a model on 
multiple tasks simultaneously.
\citet{decanlp} train a question-answering language 
model on many different NLP tasks 
(e.g., machine translation, summarization, 
sentiment analysis, etc.) by formulating 
every task as question answering
in order to make the input and output of the model be consistent.
This model can then be used to perform multiple 
tasks by asking different questions at test time.
While their overall multitask results are still below 
task-specific models, it represents a prototype for models of general 
linguistic intelligence.

In order to make progress toward models that manifest general linguistic
intelligence, we argue for 
an evaluation paradigm that rewards abilities to
(i)~deal with the full complexity of natural 
language across a variety of tasks, 
(ii)~effectively store and reuse
representations, combinatorial modules 
(e.g, which compose words into 
representations of phrases, sentences, and documents), 
and previously acquired linguistic knowledge 
to avoid \emph{catastrophic forgetting} \citep{catastrophic,french},
and (iii)~adapt to new linguistic tasks in new environments 
with little experience (i.e., robustness to domain shifts).
This is inspired by the evolving artificial general intelligence 
task suites that are used to 
develop generic exploration, planning, and reasoning abilities 
in agents that learn to interact with simulated visual environments \citep{dmlab,openaigym}. 
Here, we focus on language processing modules 
that learn
using primarily supervised and unsupervised methods 
from static corpora. We leave extensions to learning 
through interactions with simulated 
linguistic environments to future work.

In this paper we analyze the
capabilities of previously proposed algorithms as models of general
linguistic intelligence.
To directly quantify how quickly an existing model adapts to a new task we propose a new evaluation metric (\S{\ref{sec:code}}) based on
online (prequential) coding \citep{onlinecode}.
This evaluation considers the number of task-specific 
training examples needed to reach high performance, thus
rewarding sample efficiency.
Our experiments show that existing models still 
require considerable fine tuning
on each task before they perform well (\S{\ref{sec:unsuppre}}),
and although they significantly benefit from transfer 
learning on other datasets from related tasks (\S{\ref{sec:beyond}}),
they generalize poorly
to different datasets from the same task without fine tuning (\S{\ref{sec:generalization}}),
and suffer from catastrophic forgetting 
when trained on multiple datasets and tasks 
in a continual learning setup (\S{\ref{sec:transfervsmultitask}}).
We discuss the implications of our findings and 
conclude by outlining potential directions toward
general linguistic intelligence models (\S{\ref{sec:future}}).


\section{Tasks}
\label{sec:tasks}
In this section, we describe the tasks and datasets that we consider
in our experiments.
We evaluate many of our models on two main tasks: reading comprehension
on SQuAD 1.1\footnote{\url{https://rajpurkar.github.io/SQuAD-explorer/}} 
\citep{squad} and Multi Genre Natural Language Inference (MNLI\footnote{\url{https://www.nyu.edu/projects/bowman/multinli/}}; Williams et al., 2018). \nocite{mnli}
However, we also use the following tasks and datasets for training
and evaluation (e.g., in a multitask setup, for pretraining,
 and for evaluation on out-of-domain genre).

\paragraph{Reading Comprehension.}
We use two additional reading comprehension datasets:
TriviaQA\footnote{\url{http://nlp.cs.washington.edu/triviaqa/}} \citep{triviaqa}
and QuAC\footnote{\url{http://quac.ai/}} \citep{quac} 
in our experiments. While SQuAD 1.1, TriviaQA, and QuAC 
are from the same task, they have
different characteristics. 
SQuAD1.1 is a reading comprehension dataset constructed from
Wikipedia articles.
It includes almost 90,000 training examples 
and 10,000 validation examples.
TriviaQA is a dataset with question-answer pairs written 
by trivia enthusiasts and evidence collected retrospectively 
from Wikipedia and the web.
We use the Web section of TriviaQA that contains 76,000 
training examples and 300 verified validation examples.
QuAC is an information-seeking dialog-style dataset where 
a student asks questions about a Wikipedia article 
and a teacher answers with a short excerpt from the article.
It has 80,000 training examples and approximately 7,000 validation
examples.

\paragraph{Semantic Role Labeling.}
We consider a question-answer driven semantic role labeling task from \citet{qasrl}, where semantic role labeling is 
presented as a span prediction problem. Given a sentence and 
a predicate-driven wh-question, the goal 
is to predict the start and end indices 
of the answer in the sentence. 
There are around 200,000 training questions 
and 25,000 test questions in the dataset.\footnote{\url{http://qasrl.org/}}

\paragraph{Relation Extraction.}
Relation extraction is the task of extracting factual knowledge
from a corpus. \citet{uwre} show that it can be formulated as
a question answering problem by associating questions with relation slots.
We use their zero-shot relation extraction 
dataset\footnote{\url{http://nlp.cs.washington.edu/zeroshot/}}
which consists of over 900,000 training examples and almost 5,000 
test examples.

\paragraph{Natural Language Inference}
Natural language inference is a sentence pair classification 
problem where the goal is to predict 
relationships between two sentences from 
\textsc{entail}, \textsc{contradict}, and \textsc{neutral}. 
In addition to the MNLI dataset, 
we also use the 
Stanford Natural Language Inference (SNLI\footnote{\url{https://nlp.stanford.edu/projects/snli/}}; Bowman et al., 2015) dataset. \nocite{snli}
MNLI (SNLI) contains 400,000 (550,000) 
training pairs and 20,000 (10,000) test pairs respectively.

\section{Models}
\label{sec:models}
We focus on two classes of models:
self-attention models (i.e., Transformer; Vaswani et al., 2017) \nocite{transformer} and recurrent neural networks.

For self-attention models, we start from a pretrained BERT model \citep{bert},\footnote{\url{https://github.com/google-research/bert}}
which has been shown to be state-of-the-art on many
natural language processing tasks.
BERT is a big Transformer model \citep{transformer}
which is trained on two unsupervised tasks:
masked language modeling and next sentence prediction.
\citet{bert} also show that these unsupervised pretraining tasks
result in better models compared to 
the standard language modeling task (i.e., next
word prediction).
We use the BERT$_{\textsc{base}}$ model 
which has 12 Transformer layers, 
12 self-attention heads, and 768 hidden dimensions.
There are 110 million parameters in total 
in this pretrained model. 
For each task that we consider, following the original
BERT model, we add an additional
task-specific final layer to output relevant predictions.
Note that the final layer is task specific, as opposed to dataset
specific.
For BERT, we leave investigations of the ability to learn 
to understand new words to future work 
and use the default BERT vocabulary in our experiments.
We denote this model BERT.

For recurrent neural networks, 
we augment a pretrained ELMo model---which has almost 100 million parameters---with a 300-dimensional bidirectional LSTM \citep{lstm}.
We then use a bidirectional attention flow network
(BiDAF; Seo et al., 2017) \nocite{bidaf} to aggregate 
context (premise) and question (hypothesis) 
representations to predict an answer span (a label) in question 
answering (natural language inference).
We set the dimensions of the 
bidirectional LSTMs in BiDAF's modeling layer 
to 300 and output layer to 100.
Unlike BERT, ELMo is a character-based model that is able to
learn representations of new words.
We refer to this model as ELMo in our experiments.

Our hyperparameters for both models are batch size, learning rate, 
dropout rate, and $\ell_2$ regularization constant.\ignore{\cjd{Ideally we should say something about the vocabulary representation. One aspect of linguistic intelligence is the ability to learn to understand new words; I think we should be explicit about the fact that we are excluding this from our simplified model.}
\dycomment{added two sentences in the two paragraphs above (second to last sentences).
ELMo is char based, BERT is not}}

\section{Evaluating Transfer}
\label{sec:code}

Existing NLP models are typically evaluated in terms of on their performance, at the end of training, on the task of interest as measured by the performance on a held-out test set. Optimization of these metrics over years is important to
drive ongoing progress on a task. 
Aggregates of these metrics have also been used to measure (and drive)
performance of models on multiple tasks. For example,
the decaScore \citep{decanlp} is a simple summation 
of various metrics such as 
classification accuracy, $F_1$ score, \textsc{ROUGE} score, and others 
across a number of tasks.
These performance metrics capture an essential aspect of intelligence: being able to generalize from experience with a class of inputs to new inputs. However, they are incomplete as none of them asses a definining attribute of general linguistic intelligence models: 
the ability to generalize rapidly to a new task by using previously acquired knowledge.
For example, a model that only requires one 
hundred in-domain training examples 
to achieve 90\% accuracy and does not improve again with more training
examples from the same domain
arguably exhibits more desirable learning capabilities
than a model that takes one million examples 
to get to 90\% before plateauing at 92\%.

To quantify the difference in such learning behavior, we use a new online (prequential) code \citep{onlinecode} 
to evaluate model performance in our experiments, in addition to traditional performance metrics.
The online code is rooted in information theory and 
is based on connections between generalization, compression, and comprehension~\citep{wolff:1982,chaitin:2007}.
In general, the separation between training 
and test sets is arbitrary,
and there is only one set of examples for a task that 
we wish to learn rapidly.
Denote the set of $N$ examples 
$\mathcal{A} = \{(x_i, y_i)\}_{i=1}^N$ 
and the model parameters $\vect{W}$.
Consider different splits of the data 
$\mathcal{A}_1, \mathcal{A}_2, \ldots, \mathcal{A}_N$, where
$\mathcal{A}_i$ is a subset that contains 
$\{(x_j, y_j)\}_{j=1}^{i}$ for
some ordering of the examples (i.e., $\mathcal{A}_i$
is the subset that contains $i$ examples
from $(x_{1}, y_{1})$ 
to $(x_{i}, y_{i})$).
Denote parameters of the model trained on a particular 
subset as $\vect{\hat{W}}_{\mathcal{A}_i}$.

We consider an online setup
where the codelength is computed as:
\begin{align}
\label{eq:online}
\ell(\mathcal{A}) = \log_2 |\mathcal{Y}| - \sum_{i=2}^N \log_2 p(y_{i} \mid x_{i}; \vect{\hat{W}}_{\mathcal{A}_{i-1}}),
\end{align}
where $|\mathcal{Y}|$ is the number of possible classes (labels) in the data.

The codelength $\ell(\mathcal{A})$ can be interpreted as follows. Assume Alice has all $(x_i,y_i)$ pairs in $\mathcal{A}$, Bob has just the $x_i$'s from $\mathcal{A}$, and that Alice wishes to communicate the $y_i$'s to Bob. One procedure would be for Alice to train a neural network, and send its parameters to Bob, but neural networks have larges numbers of parameters that are expensive to communicate. The online procedure solves this problem by having Alice and Bob agree on the form of the model, its initial random seeds, and its learning algorithm. Alice starts by communicating $y_1$ with a uniform code, and then both Alice and Bob learn a model that predicts $y$ from $x$, and Alice uses that model to communicate $y_2$. The process repeats where both parties learn from ever larger datasets; since both models are trained using the same procedure, they stay in sync. This process continues until the entire dataset has been transmitted. In this sequential evaluation, 
a model that performs well with a limited number
of training examples will be rewarded by having 
a shorter codelength (Alice will require fewer bits to transmit the subsequent $y_i$ to Bob). 

\ignore{
Next, consider the case where there are separate training and test
sets. Consider a simpler setup where $\mathcal{A} \neq \mathcal{B}$ 
but $M = N$.
We compute the ``online'' codelength of the test data as:
\dycomment{so this is an unusual setup in the
information theoretic perspective.
it's like alice and bob only know the labels for their training data
at certain timesteps, but alice needs to keep sending bob
test labels everytime a new subset of training labels are
revealed. but why do bob and alice not train the model
on the newly transmitted test labels to do even better?
so in this setup both bob and alice cannot use the test
labels information to train.
if we don't care about the information theory
aspect of the metric, why don't we always 
evaluate on all test data
but with increasing number of training data?}
\begin{align}
\label{eq:preq}
\ell(\mathcal{B}) = - \sum_{i=1}^{N} \log_2
p(y^{\text{test}}_i \mid x^{\text{test}}_i; \vect{\hat{W}}_{\mathcal{A}_i}).
\end{align}
Intuitively, we evaluate the model on a new test example
after we see an additional example of the training set.
As a result, a model that performs well with a limited number
of training examples will have a lower online codelength.
}

In the above formulation, the model is 
evaluated for every example in the training data, 
which can be very expensive in practice.
A more realistic approach is to split 
$\mathcal{A}$ into $M$ increasing subsets, $\mathcal{S}_1,\mathcal{S}_2,\ldots,\mathcal{S}_M$ where $\mathcal{S}_M=\mathcal{A}$ and 
only evaluate the model $M$ times.
In this case, we have:
\begin{align*}
\ell(\mathcal{A}) = |\mathcal{S}_1| \log_2 |\mathcal{Y}| - \sum_{i=2}^M \log_2 p(y_{\mathcal{S}_{i}} \mid x_{\mathcal{S}_{i}}; \vect{\hat{W}}_{\mathcal{S}_{i-1}}),
\end{align*}
where $\mathcal{S}_i$ denotes the $i$-th data subset.

Online codelength is related to area under the learning curve~\citep{guyon:2011},
as they share similar motivations, and also to dynamic sequence model evaluation~\citep{krause:2018}.
Like our online codelength, both also require a predefined dataset ordering.
We note that while all these metrics consider 
the number of examples required to
achieve a certain performance, they still do not fully
consider model complexity 
(e.g., the number of parameters in the model)
and training cost 
(e.g., how much computational resources are needed,
how many hours, etc.).
However, three nice features of online codelength 
are that: (i)~it can be used 
across a number of tasks by any \emph{probabilistic} model,
(ii)~it allows seamless incorporation of other model and training 
properties (e.g., model complexity, training cost) if desired, 
and (iii)~it will reflect improvements in generalization performance 
due to better architecture, better initialization, 
and better learning algorithms.
In our experiments below, we also show that it
correlates well with standard evaluation metrics 
such as accuracy and $F_1$ scores.

\section{Experiments}
\label{sec:exp}

\subsection{Unsupervised Pretraining}
\label{sec:unsuppre}
\emph{How much in-domain training data is needed to obtain good performance?\\}

\begin{figure}[t]
    \centering
    \includegraphics[scale=0.35]{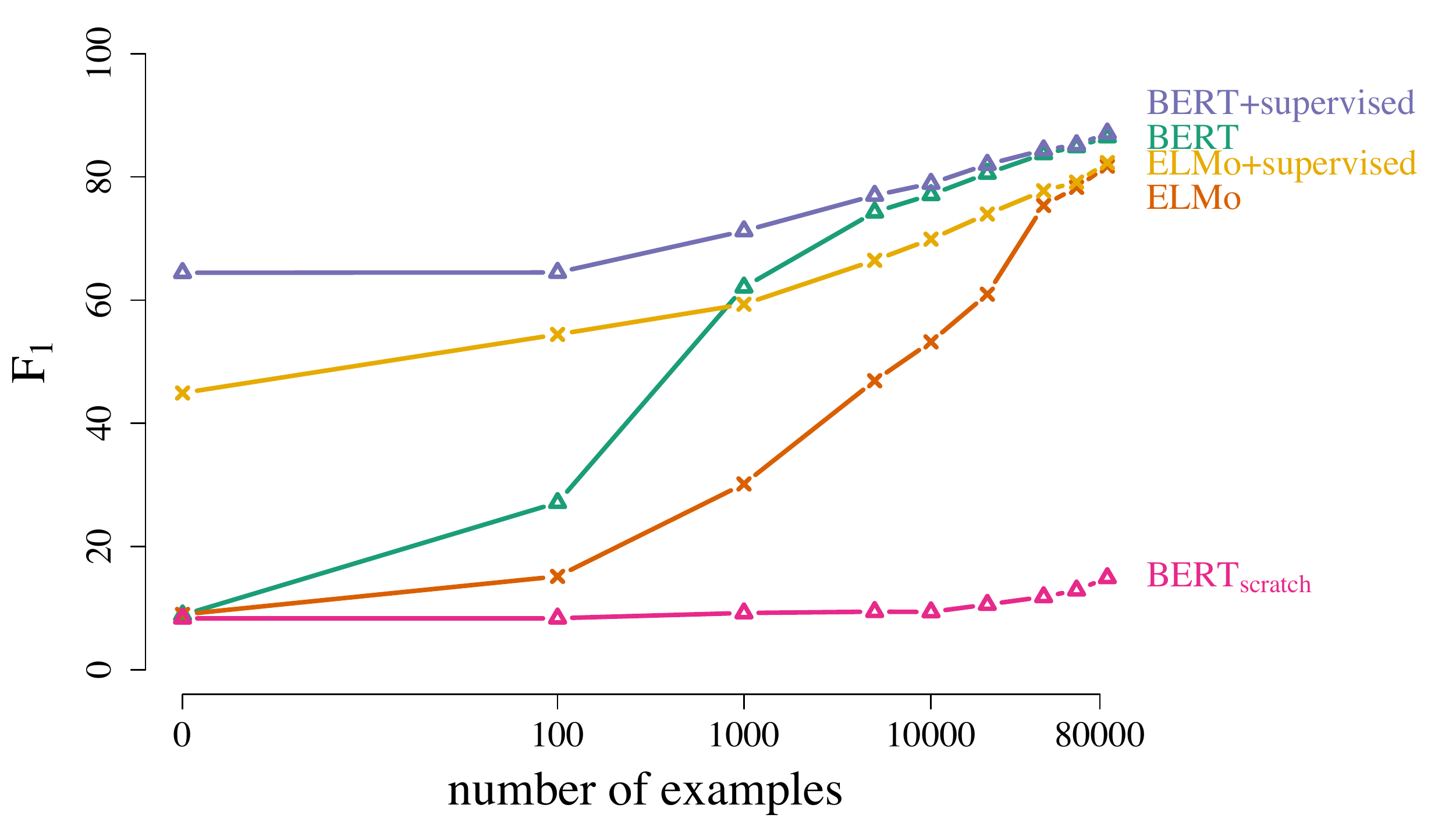}
    \caption{$F_1$ scores on SQuAD as a function of the number of training examples (log scale). BERT+supervised and ELMo+supervised denote BERT and ELMo models that are pretrained on other datasets and tasks (see \S{\ref{sec:beyond}}), $\text{BERT}_{\text{scratch}}$ denotes a Transformer with a similar architecture to BERT that is not pretrained on any unsupervised task at all (i.e., trained from scratch).}
    \label{fig:unsup1}
\end{figure}

\begin{figure}[t]
    \centering
    \includegraphics[scale=0.35]{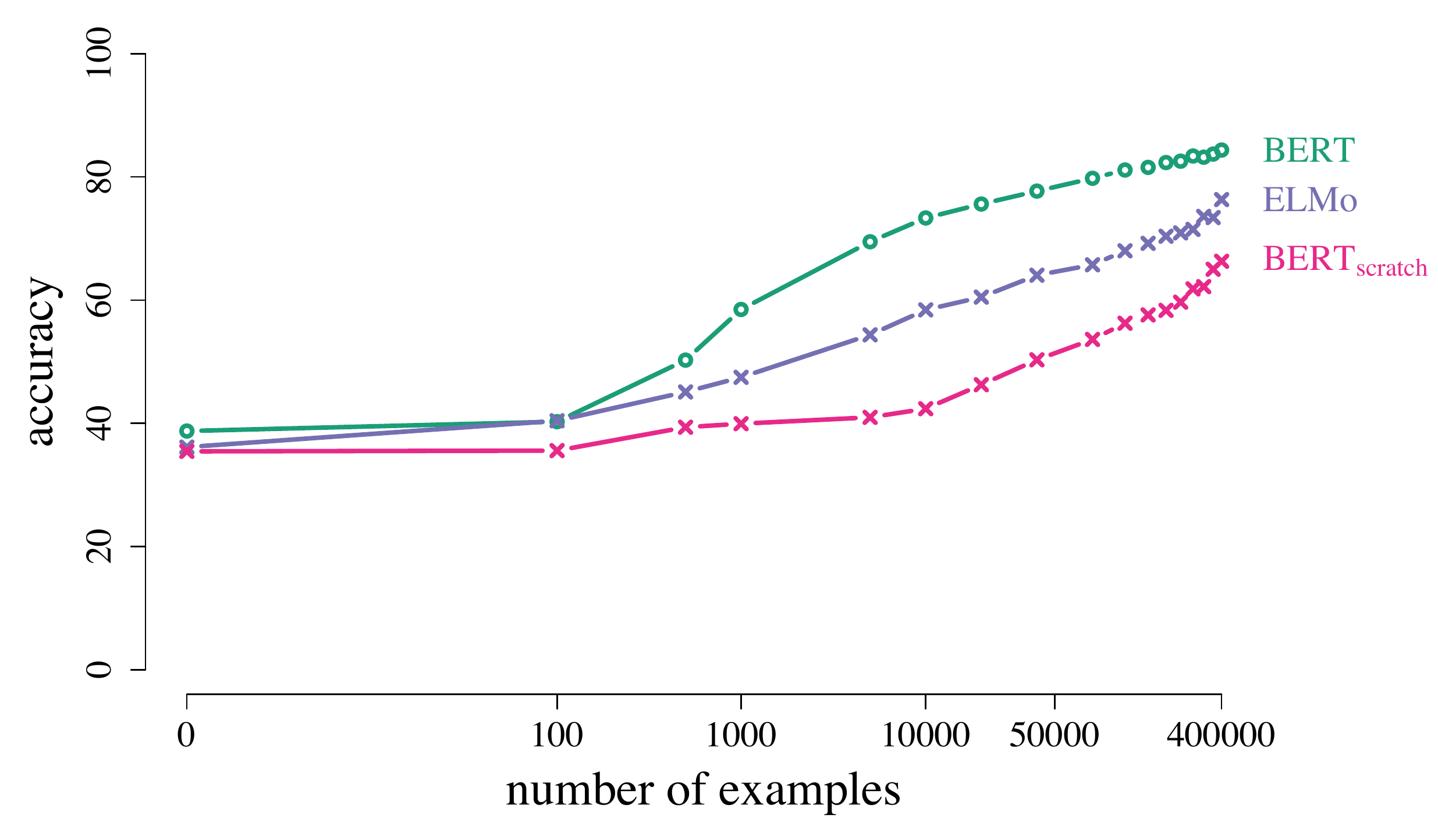}
    \caption{Classification accuracy on MNLI as a function of the number of training examples (log scale). $\text{BERT}_{\text{scratch}}$ denotes a Transformer with a similar architecture to BERT that is not pretrained on any unsupervised task at all (i.e., trained from scratch).}
    \label{fig:unsup2}
\end{figure}

Our first set of experiments is designed to investigate 
how much training data is needed to achieve
high performance on two language understanding tasks:
reading comprehension on SQuAD 1.1 and natural language inference on MNLI.
We show learning curves of two models as a function
of the number of training examples on SQuAD1.1 and MNLI validation sets 
in Figure~\ref{fig:unsup1} and Figure~\ref{fig:unsup2}.
On both SQuAD and MNLI, both models are able to approach their 
asymptotic errors after seeing approximately
40,000 training examples, a surprisingly 
high number of examples
for models that rely on pretrained modules 
(i.e., BERT and ELMo).
While adding more training
examples improves both models, the performance gains
are not as significant as the first 40,000 examples.
Between these two models, BERT outperforms ELMo 
on both datasets and approaches its asymptotic error faster.
Previous work have established the benefits
of transfer learning from unsupervised tasks 
both in terms of sample complexity and overall performance 
\citep{daile,openai,bert}.
We also verify this in our experiments by training a Transformer
with the same architecture as BERT on SQuAD and MNLI from
scratch on the entire training set.
This model can only achieve 14.9 $F_1$ 
score on SQuAD and 66.3 accuracy on MNLI
after 500,000 training iterations, 
far below the performance
of a similar model pretrained on unsupervised tasks.
We note that our $F_1$, exact match, and accuracy scores 
on SQuAD and MNLI validation sets are
upper bound best results since we tune hyperparameters on 
the respective validation sets. 

We next split the original SQuAD training set 
into 8 example subsets of 10,000 examples each 
and MNLI into 10 example subsets of 40,000 examples.
We use uniform encoding for predicting 
the first example subset,\footnote{We observe that
using uniform encoding for the first subset is better than using 
predictions from pretrained models with randomly initialized final layers.}
use a ``default'' hyperparameter configuration 
for training the first model to predict the second example subset
(i.e., we choose reasonable hyperparemeter values based on intuitions),
and select hyperparameters of the best model
from the previous subset for subsequent subsets 
to avoid tuning on evaluation subsets for predictions.
We obtain online codelengths 
of 102.42 kbits (BERT) and 112.96 kbits (ELMo) for SQuAD1.1 
and 89.25 kbits (BERT) and 132.17 kbits (ELMo) for MNLI.
We can see that codelengths correlate with other evaluation metrics
such as $F_1$, exact match, and accuracy on these datasets (lower
codelengths generally means higher $F_1$, exact match, or accuracy),
consistent with the findings of \citep{onlinecode}.
We use these as baseline codelengths for our next set of
experiments in \S\ref{sec:beyond}.

\subsection{Beyond Unsupervised Pretraining}
\label{sec:beyond}
\emph{Can pretraining on other datasets and tasks improve performance?\\}
We next investigate a more general transfer learning
strategy using related tasks beyond unsupervised objectives.
\citet{bert} show that jointly training BERT on SQuAD and TriviaQA
slightly improves final performance.
Here, we focus on a pretraining setup to highlight the benefit of transfer. 
We discuss joint (multitask) training in \S{\ref{sec:transfervsmultitask}}.

We pretrain BERT and ELMo 
models on semantic role labeling, 
relation extraction, natural language inference, and other reading comprehension datasets (TriviaQA and QuAC) described in \S\ref{sec:tasks}.
We randomly sample a batch of examples from one of the tasks above in
the pretraining phase for 100,000 iterations.
We then take these pretrained models and train on only SQuAD.
Table~\ref{tbl:supresults} show overall validation set 
results on SQuAD for each model.

\begin{table}[h]
    \centering
    \begin{tabular}{|l||r|r|r|}
    \hline
    \textbf{Model} & \textbf{EM} ($\uparrow$) & \textbf{$F_1$} ($\uparrow$) & \textbf{codelength} ($\downarrow$)\\
    \hline
     \textsc{BERT} & 78.5 & 86.5 & 102.4\\
     \textsc{BERT} + supervised & 79.4 & 87.1 & 31.7\\
     \textsc{ELMo} & 72.1 & 81.8 & 113.0\\
     \textsc{ELMo} + supervised & 72.8 & 82.3 & 54.5\\
    \hline
    \end{tabular}
    \caption{Results on SQuAD for BERT and ELMo pretrained on other supervised tasks vs. from only unsupervised tasks}
    \label{tbl:supresults}
\end{table}

The results show that pretraining on other datasets and tasks
slightly improve performance in terms of final exact match
and $F_1$ score.
However, in terms of codelength, the models pretrained on other
tasks significantly outperform the non-pretrained ones.
A major reason for this is that pretraining on other
question answering datasets trains the final
layer (i.e., the decoder) used in SQuAD (since many of the 
pretrained datasets are question answering datasets), 
so the first example subset when computing the codelength 
can be predicted without using the uniform encoding.
It is analogous to learning what a
question answering task is, even though they are
learning from different datasets.

\begin{figure}[t]
    \centering
    \includegraphics[scale=0.35]{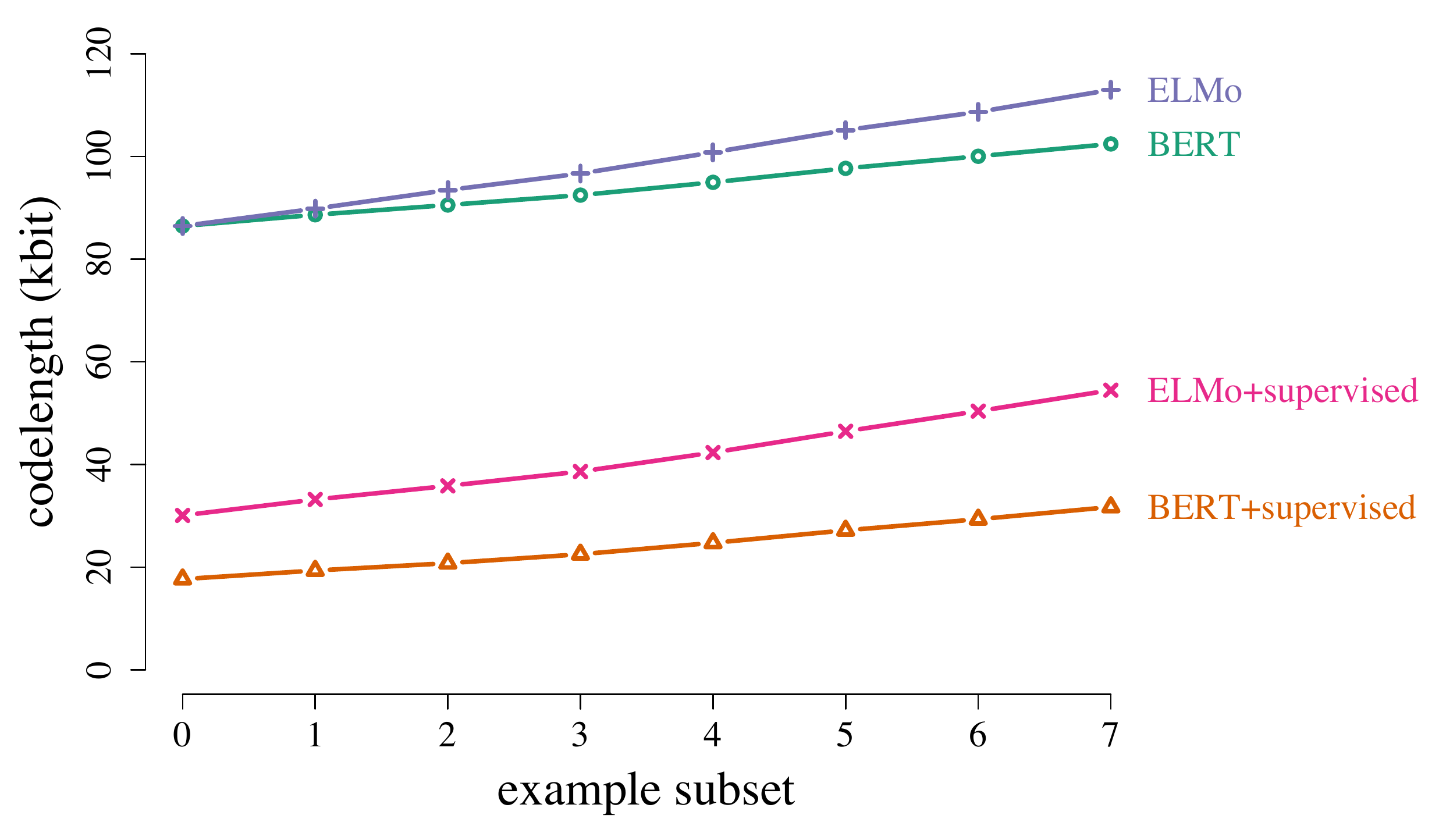}
    \caption{Online codelengths for different example subsets on SQuAD for BERT and ELMo pretrained on other supervised tasks vs. from only unsupervised tasks.}
    \label{fig:supcode}
\end{figure}

As shown in Figure~\ref{fig:supcode}, the models
are able to predict answers reasonably well even 
before seeing any SQuAD-specific training examples.
BERT and ELMo  achieve 62.9 and 43.3 $F_1$ scores 
after training on other tasks and datasets before
seeing any SQuAD examples.
These results give rise to an 
interesting question about how well models that
are trained only on SQuAD perform on these other
datasets, which we answer in \S{\ref{sec:generalization}}.
This experiment also shows that an online 
evaluation paradigm such as online codelength
highlights different model characteristics 
that are not captured by metrics such as $F_1$ score
or accuracy.
A characteristic of the codelength metric is that 
models that perform significantly worse at the beginning 
can have problems catching up to the performance of 
models that start well, often referred to as the 
catch-up phenomenon \citep{catchup}. 
If such a property is not desired, we can switch between models by
always choosing the best model to encode a particular subset
when computing the codelength (we do not do this in our experiments).

Since we observe benefits of transfer learning from other tasks,
another important question is how to select tasks to pretrain on,
and whether there is a training curriculum that a model should follow.
Ideally, a general linguistic intelligence 
model should be able to consume examples from any kind of task
in any order.
Our experiments in \S{\ref{sec:transfervsmultitask}} will provide
some insights into training curriculum.

\subsection{Generalization}
\label{sec:generalization}
\emph{Do these models generalize to other datasets from the same task?\\}

Our next set of experiments is to analyze 
generalization properties of existing models.
We investigate whether our models overfit
to a specific dataset (i.e., solving the dataset) it is trained on or 
whether they 
are able to learn general representations and modules (i.e., solving the task).

We take our best SQuAD models from \S\ref{sec:unsuppre}
and evaluate them on other question-answering datasets to see
whether these models generalize to other datasets 
from the same task.
Table~\ref{tbl:generalization} shows results on
TriviaQA, QuAC, QA-SRL, and QA-ZRE.\footnote{Our results
on these four datasets are not necessarily comparable with
results from other work since we remove unanswerable questions (as identified by the dataset creator) for simplicity.}
We see that high-performing SQuAD models 
do not perform well on other datasets without 
being provided with training examples from these datasets.
These results are not surprising since the examples come
from different distributions, but they do highlight the fact that there is a substantial gap between learning a task and learning a dataset.

Our results indicate
that models that work well on SQuAD still require
training examples from other datasets 
before it can serve as a general-purpose question answering model.
However, our previous experiments in \S{\ref{sec:beyond}}
show that training on other tasks and datasets 
speed up learning on SQuAD greatly, 
so there is useful information that can be 
extracted from these tasks.
We next discuss the interaction 
between training curriculum and model performance.

\begin{table}[h]
    \centering
    \begin{tabular}{|l|r|r|r|r|r|}
    \hline
    \textbf{} & \textbf{SQuAD} & \textbf{Trivia} & \textbf{QuAC} & \textbf{QA-SRL} & \textbf{QA-ZRE}\\
    \hline
    BERT & 86.5 (78.5) & 35.6 (13.4) & 56.2 (43.9) & 77.5 (65.0) & 55.3 (40.0)\\
    ELMo & 81.8 (72.2) & 32.9 (12.6) & 45.0 (34.5) & 68.7 (52.3) & 60.2 (42.0)\\
    \hline
    \end{tabular}
    \caption{$F_1$ (exact match) scores of the best BERT and ELMo models trained on SQuAD and evaluated on other question answering datasets.}
    \label{tbl:generalization}
\end{table}

\begin{figure}[t]
    \centering
    \includegraphics[scale=0.35]{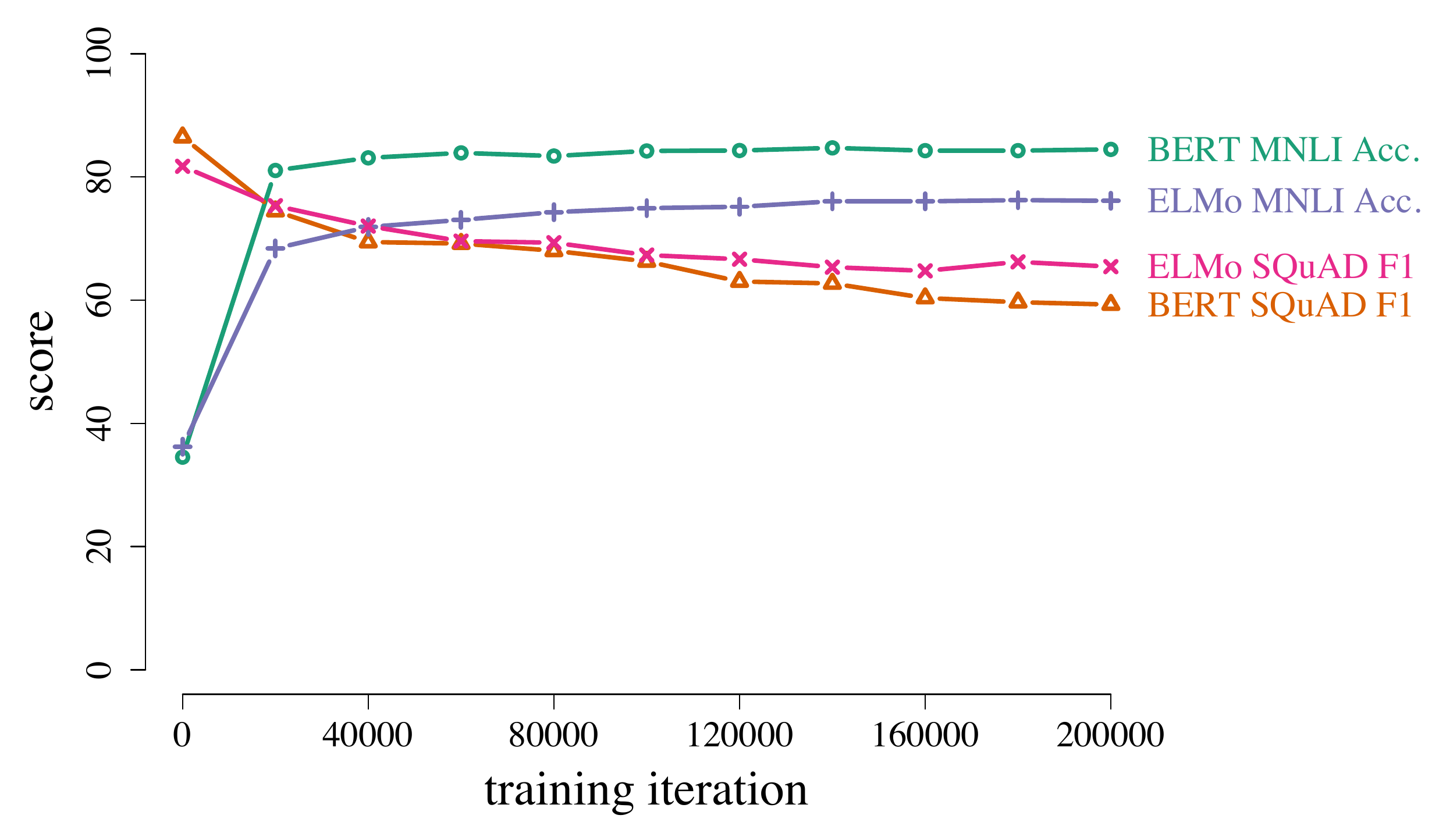}
    \includegraphics[scale=0.35]{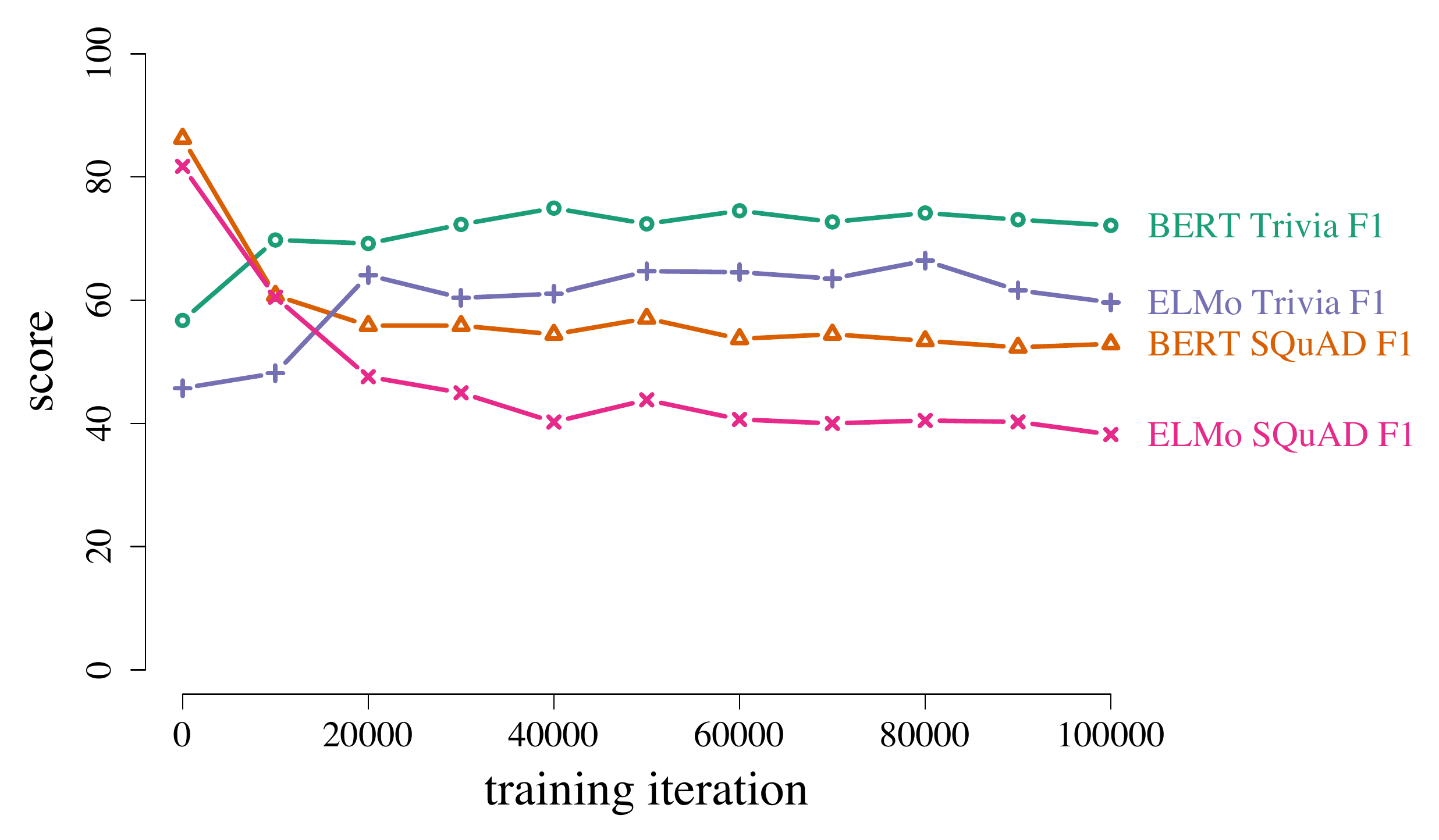}
    \caption{Catastrophic forgetting in a continual learning setup on
    unsupervised $\xrightarrow{}$ SQuAD $\xrightarrow{}$ MNLI (top) and
    unsupervised $\xrightarrow{}$ SQuAD $\xrightarrow{}$ TriviaQA (bottom).}
    \label{fig:catastrophic}
\end{figure}

\begin{figure}[t]
\centering
\includegraphics[scale=0.35]{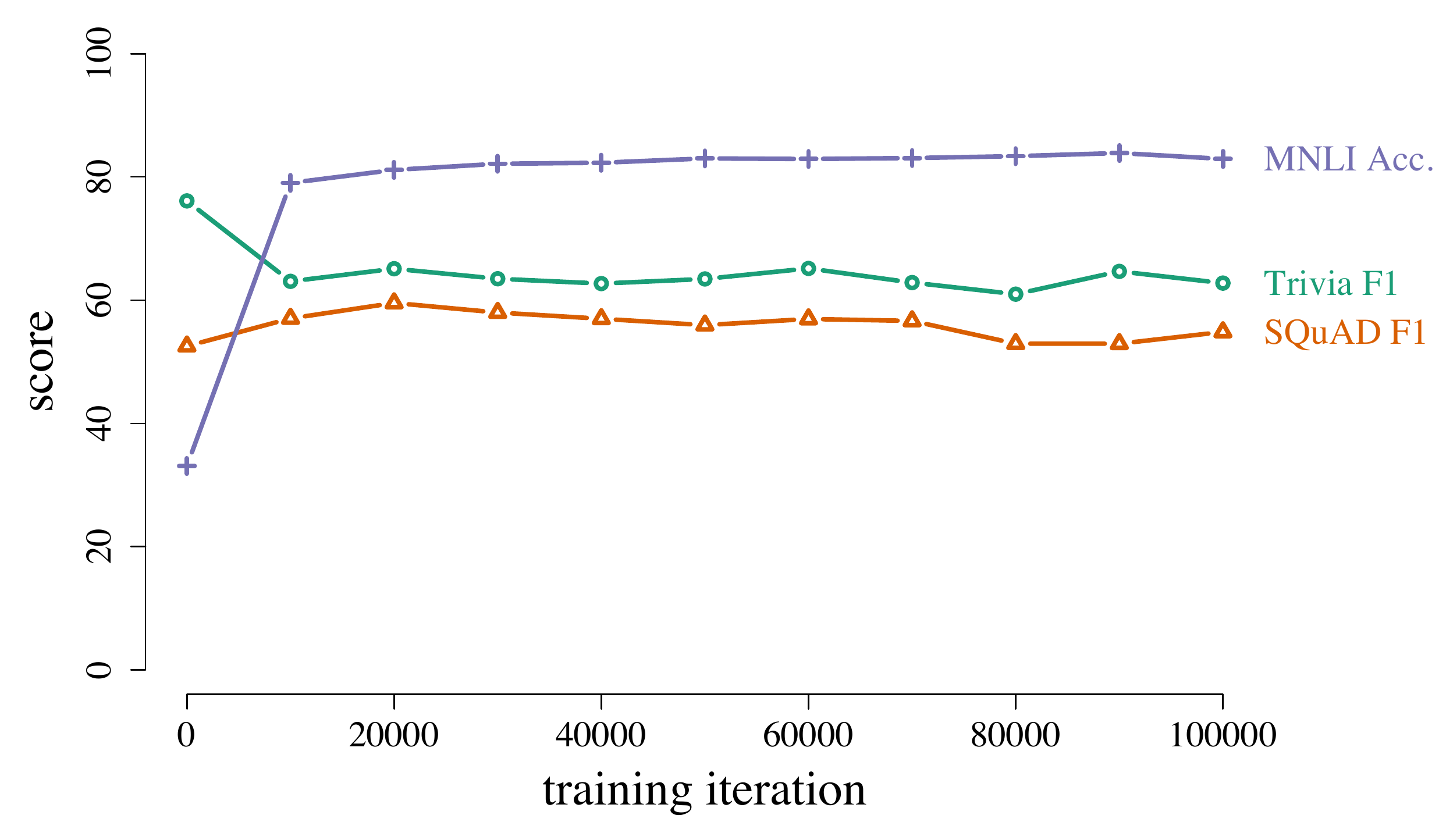}
\caption{Catastrophic forgetting in a continual learning setup for BERT on
    unsupervised $\xrightarrow{}$ SQuAD $\xrightarrow{}$ TriviaQA $\xrightarrow{}$ MNLI (bottom).}
\label{fig:catastrophic2}
\end{figure}

\subsection{Curriculum and Catastrophic Forgetting}
\label{sec:transfervsmultitask}
\emph{How fast do these models forget their previously acquired linguistic knowledge?\\}
\emph{How does curriculum affect performance and how do we design this curriculum?\\}

An important characteristic of general linguistic intelligence models
is the ability to store and reuse linguistic knowledge throughout their
lifetime and avoid catastrophic forgetting.
First, we consider a continual learning setup, where we train our best SQuAD-trained 
BERT and ELMo from \S{\ref{sec:unsuppre}} on a new task, TriviaQA or MNLI.
Figure~\ref{fig:catastrophic} shows results of our experiments.
We can see that in a continual learning (transfer) setup, the performance on both models on the first supervised task they are trained on (SQuAD)
rapidly degrades. This is true even for MNLI, which has  
a task-specific final layer for classification that is 
not shared with the original question-answering tasks (thus indicating that the underlying representations being computed are changing).
We search over a wide range of possible learning rate initial values 
and none of the models that obtain good results on the new
task are able to maintain performance on SQuAD.
For BERT, we also include an extended continual learning result for: unsupervised $\xrightarrow{}$ SQuAD $\xrightarrow{}$ 
TriviaQA $\xrightarrow{}$ MNLI in Figure~\ref{fig:catastrophic2} (using the model from unsupervised $\xrightarrow{}$ SQuAD $\xrightarrow{}$ 
TriviaQA in the previous experiment). 
Interestingly, adding a new MNLI task does not decrease 
the performance on SQuAD further, although it does make the model
forget how to do well on TriviaQA.

\begin{figure}[t]
    \centering
    \includegraphics[scale=0.35]{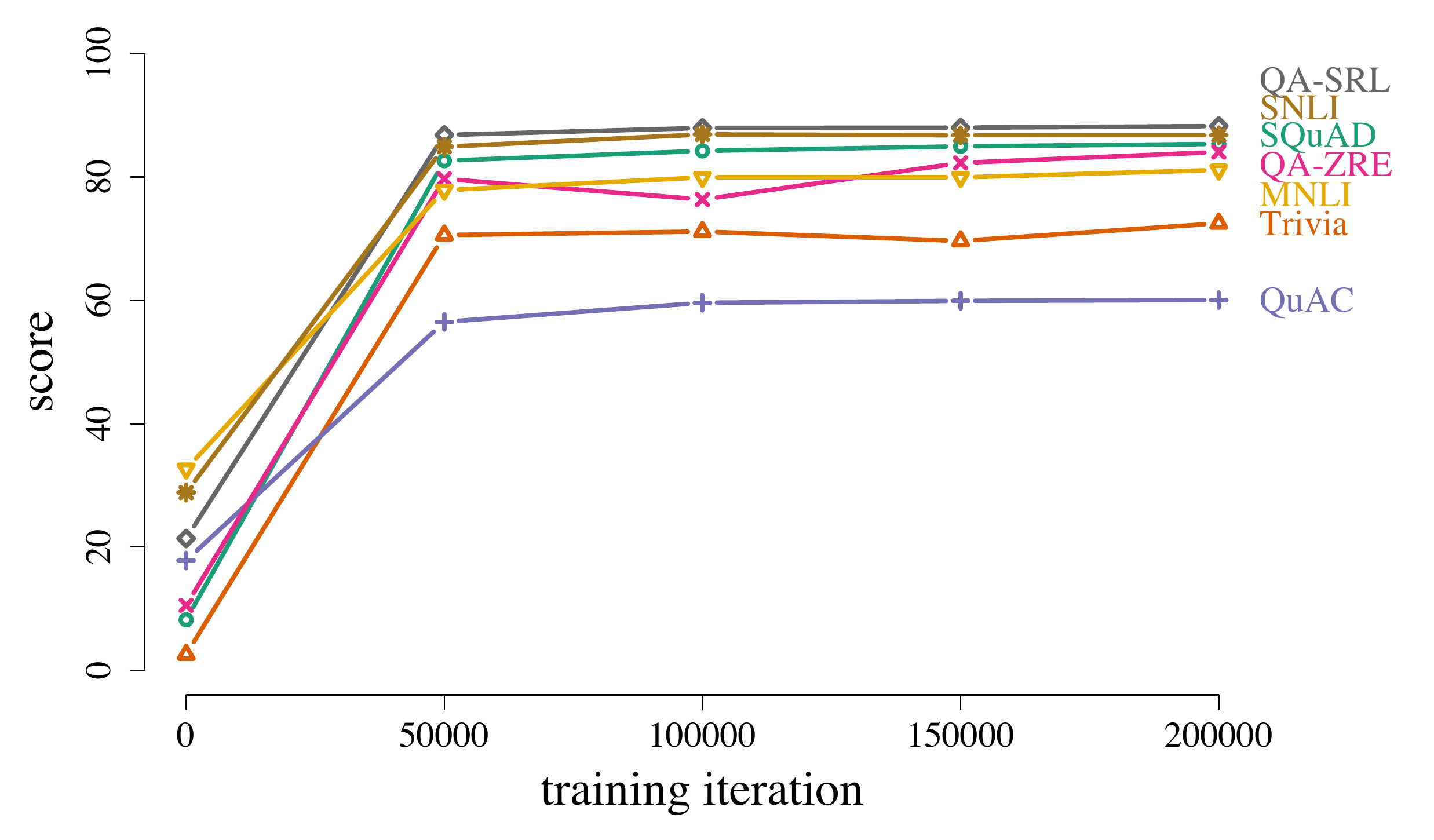}
    \caption{Model scores on various datasets with a random training curriculum.}
    \label{fig:multitask}
\end{figure}

We next investigate whether there is a curriculum 
that allows these models to perform well on all tasks.
We train BERT and ELMo on all tasks in \S{\ref{sec:tasks}},
where at each iteration we sample a batch of examples from 
a randomly chosen task (with uniform probability).
Table~\ref{tbl:multitaskresults} shows performance of these models
after 200,000 training iterations. Interestingly, unlike in the
continual learning setup, these models
are able to perform reasonably well on many tasks (especially BERT).
Figure~\ref{fig:multitask} shows performance of BERT on
all tasks as training progresses. We can see that using a random training
curriculum allows the model to not forget previously acquired knowledge,
and that after about 50,000 iterations the model can work reasonably well on all tasks.
However, the drawback of such a curruculum is that 
it either requires all tasks to be presented at the beginning or
retraining on all tasks after seeing a new task.
It is therefore still an open question how to transfer effectively in a
continual learning setup.

\begin{table}[h]
    \centering
    \begin{tabular}{|l|r|r|r|r|r|r|r|}
    \hline
    \textbf{} & \textbf{SQuAD} & \textbf{Trivia} & \textbf{QuAC} & \textbf{QA-SRL} & \textbf{QA-ZRE} & \textbf{MNLI} & \textbf{SNLI}\\
    \hline
    BERT & 85.4 & 72.5 & 60.0 & 85.0 & 88.2& 81.1 & 88.0\\
    ELMo & 78.3& 57.1& 54.3& 67.3& 88.5& 69.1& 77.9\\
    \hline
    \end{tabular}
    \caption{Results on multiple tasks for BERT and ELMo trained with a random training curriculum for 200,000 iterations.}
    \label{tbl:multitaskresults}
\end{table}

\section{Discussion}
\label{sec:future}
Our series of experiments show that while we are making great progress as a field, 
we are still missing several 
fundamental components to achieve
general linguistic intelligence models.
Existing state-of-the-art models are
large parametric deep learning models trained in a discriminative 
fashion on many (unsupervised and supervised) training examples.
They typically include task-specific
components to perform multiple tasks, and assume
that the data distribution is stationary within a task.
While such an approach is reasonable, 
we see that fine tuning these components still
require a fair number of training examples,
and that these models are very prone to catastrophic
forgetting. 

Our results show that we need more
sophisticated \textbf{transfer and continual 
learning methods}. For instance, techniques such as
elastic weight consolidation \citep{ewc} and
progress and compress \citep{progresscompress} may hold 
promise for improving our models' robustness to forgetting.
Another crucial component that is currently
underexplored is a \textbf{memory module} that
rapidly adapts to domain shifts and generalizes across tasks.
Recent work on neural cache \citep{grave} 
and dynamic evaluation \citep{krause:2018}
provide some evidence of the 
effectiveness of this approach
for language modeling.
Investigating how to design similar models that operate
beyond the token level and are reusable across tasks
is a promising direction.

Learning to learn (i.e., \textbf{meta learning}) is another approach to
learn models that adapt to new tasks rapidly. 
In this setup, parameters of the model are trained to ensure that
a small number of gradient steps on a new task results in
good performance on the task.
There has not been much work on
meta learning natural language processing models.
Standard meta learning methods \citep{maml,reptile} 
require the distribution
of tasks that we want the model to adapt to to be known in advance
(i.e., we need to know all the tasks that we want to evaluate on).
However, since NLP models take the same kind of inputs for all tasks 
(i.e., a sequence of characters or words),
it is conceivable that meta-learned NLP models
generalize to new unseen tasks. 
In particular, our codelength objective offers an intriguing meta learning objective since it quantifies generalization speed.

When presented with multiple training 
examples from multiple tasks,
our experiments also show that \textbf{training 
curriculum} can have a considerable
effect on model performance.
In a scenario where we can see all 
the tasks beforehand (i.e., similar to the
multitask training setup), designing a curriculum that allows the model
to learn better and faster becomes important.
Recent progress on methods such as
Population Based Training (PBT, Jaderberg et al., 2017) could be used
to design a curriculum that improves training.\nocite{pbt}
PBT requires a lot of computational resources, but the
cost could be amortized if it can be used to establish
a fixed schedule that becomes a standard curriculum for
training general linguistic intelligence models.

One key reason that our models generalize poorly to other
tasks is that they rely on task-specific components (among others).
A perfect language model 
should in theory be able to do any 
linguistic task (e.g., by formulating the task as a question
and querying the language model to generate answers).
\citet{decanlp} proposed such a model that can work on
multiple tasks, although the overall
performance of their model is still below task specific models.
Unsupervised pretraining of language models 
that are then used as a backbone of task-specific models
has also been shown to considerably improve
performance on downstream tasks \citep{openai,elmo,bert}.
Therefore, we believe that continued progress on \textbf{generative language 
models} will drive progress on general linguistic intelligence.

In this paper, we mainly consider models that learn passively 
by observing environments (i.e., from static corpora) 
and we evaluate them on sample efficiency, generalization 
to other tasks, and their ability to 
avoid catastrophic forgetting.
There are other important skills that we do not consider
such as robustness to adversarial examples 
or the ability to understand multiple languages.
\citet{adversarial} show that reading comprehension models
trained on SQuAD are very prone to adversarial attacks,
although some adversaries are constructed by shifting the distribution of the inputs in non-trivial but meaning-preserving ways.
Improving cross-dataset transfer inside a single task will likely see this classes of adversaries weakened.
Ideally, a general linguistic intelligence model should also be able 
to provide human-understandable reasoning of its decisions and intentions 
(interpretablity and explainability).
We leave investigations of such abilities for future work.

\section*{Acknowledgements}
We thank Stephen Clark for helpful comments on an earlier draft of this paper.

\bibliography{main}
\bibliographystyle{main}
\end{document}